# Least-Squares FIR Models of Low-Resolution MR data for Efficient Phase-Error Compensation with Simultaneous Artefact Removal

Joseph Suresh Paul[*], Uma Krishna Swamy Pillai, Nyjin Thomas

Indian Institute of Information Technology and Management, Trivandrum, India

[*] j.paul@iiitmk.ac.in

## Abstract

Signal space models in both phase-encode, and frequency-encode directions are presented for extrapolation of 2D partial kspace. Using the boxcar representation of low-resolution spatial data, and a geometrical representation of signal space vectors in both positive and negative phase-encode directions, a robust predictor is constructed using a series of signal space projections. Compared to some of the existing phase-correction methods that require acquisition of a pre-determined set of fractional kspace lines, the proposed predictor is found to be more efficient, due to its capability of exhibiting an equivalent degree of performance using only half the number of fractional lines. Robust filtering of noisy data is achieved using a second signal space model in the frequency-encode direction, bypassing the requirement of a prior highpass filtering operation. The signal space is constructed from Fourier Transformed samples of each row in the low-resolution image. A set of FIR filters are estimated by fitting a least squares model to this signal space. Partial kspace extrapolation using the FIR filters is shown to result in artifact-free reconstruction, particularly in respect of Gibbs ringing and streaking type artifacts.

## Introduction

Phase errors in MRI can result from Off resonance effects due to imperfections of the static magnetic field, or significant changes in the susceptibility within the imaged field-of-view (FOV) [1-4]. The latter can often result in image distortion, particularly along natural and artificial air-filled cavities. When phase errors are present, the complete kspace is no longer conjugate symmetric. Coverage of complete kspace requires acquisition of all the phase-encode and frequency-encode lines for a given resolution. Consequently, constrained image reconstruction using either half the set of phase-encode lines or the complete set of partial echoes excluding the dephasing lobe and covering the entire kspace, may result in loss of anatomical details. Thus application of phase correction methods necessitates acquisition of fractional lines either in the phase-encode direction or frequency-encode direction, and/or both. The resulting partial kspace can be either 1D, or 2D depending on the acquisition of fractional lines in any one, or both directions.

Imposition of temporally constrained acquisition, or acquisition of incomplete set of phase-encodes can result in truncation effects [5]. Alternatively, the effects can also be produced from under-sampling in either phase or frequency-encode directions. Often, these artefacts referred to as "*Gibbs Ringing*" are manifested as striations of sharper image edges in the form of repeated bands parallel to the edge [6]. Alleviation of these artefacts by windowing in the partial kspace, can lead to blurring and loss of spatial resolution. A second type of artefact arises from performing extrapolation of missing information in kspace derived from highpass filtered low SNR images [7]. This type of artefact, often manifested as streak lines, results mainly from noise amplification due to highpass filtering. The solutions for alleviation of phase errors, and simultaneous measures for artifact suppression have been broadly addressed through taxonomy of three distinct approaches. These consist of 1) methods based on low-resolution symmetric data, 2) parametric models, and 3) statistical estimation. Statistical methods use information obtained from the phase of first and higher order autocorrelation, calculated from complex valued phase-distorted image [8]. While the statistical methods are applicable only in situations where the phase variation due to field inhomogeneity is minimal, the usage of higher order autocorrelation results in phase wrapping due to large pixels shifts. The proposed method comprises of a combination of the first two approaches. This results in a more efficient phase correction, since the same performance can be achieved using a lesser number of fractional lines.

In the first group of methods, a narrow strip of data having symmetric phase-encode coverage is selected to provide a low-resolution approximation of the spatial phase variation. Phase of this low-resolution data is then used for phase correction. In the homodyne method [9], the partial kspace data is multiplied with a weighting function called merging filter. The weighted partial kspace is then phase-corrected using the low-resolution symmetric data. In the iterated version of this approach known as POCS, the missing kspace lines in each iteration are replaced from the kspace of the phase-corrected image [10]. It has been possible to reduce the blurring effects associated with conventional homodyne by applying phase-correction to the highpass filtered image [11]. Nevertheless, attempts on prior highpass filtering can result in noise amplification.

With the model-based approach, extrapolation of missing kspace data is accomplished using parametric models. A form of ARMA known as the transient error reconstruction method has been used to model MRI data successfully [12]. Unfortunately, the autoregressive portion of the ARMA model can result in poles in the transfer function, which produce high intensity spikes in the image. It would be much easier to model MRI signals if the corresponding image-domain data contained only sharp features. It is possible to transform time-domain MRI data such that Fourier transformation produces an image, which consists of predominantly sharp lines [13]. Such a transformed image can be obtained by applying a linear high-pass filter. Equivalently, representation of the low-resolution image as a series of boxcar functions [14], leads to the Fourier Transform of the spatial derivatives modeled as a summation of complex sinusoids in noise. This framework entails the application of linear prediction for extrapolation of partial kspace samples weighted using the appropriate frequency terms [15]. Increasing the efficiency of phase correction using a predictor along phase-encode direction, in the frequency-weighted kspace followed by homodyne correction is a possible solution. The predictor may result in unstable filter weights, leading to spurious spikes, or high intensity streaks in the final reconstruction.

In view of these difficulties, we propose a form of signal space representation of kspace data. The collection of instantaneous kspace samples over a given range of phase-encode values is represented as a signal vector. The evolution of signal vectors in both positive and negative phase-encode directions is illustrated in Fig. 1a. Starting with the signal vector representative of the kspace sample at a given time point on the zero phase-encode line, inclusion of successive samples corresponding to each incremental step of phase-encoding rotates the signal vector in counter-clockwise direction for positive phase-encodes, and clockwise direction for negative phase-encodes. The presence of noise is shown to generate an additional offset $\theta$ between the directions of the two resultant signal vectors, as shown in Fig. 1b.

Fig. 1 Here

In order to fit the signal space model with the linear prediction formulation of [13-14], it would be ideal to consider the signal spaces using samples from the Fourier Transformed spatial derivative. In an effort to increase the number of fractional lines for efficient phase correction, a linear predictor, may therefore be applied to extrapolate kspace samples along negative phase-encode direction in this filtered transform domain. However, a direct application of the predictor can result in large variance of the predicted samples, particularly for higher phase-encodes. As a countermeasure, we resort to apply a one-step predictor in an iterative manner. In this process, the predicted sample is included in the data set used for performing prediction in the each successive iteration. This can lead to accumulation of errors causing artifacts in the image reconstruction. In this paper, we develop a subspace projection based approach to compensate for such errors. However, the subspace projection works only when the angle θ between the resultant signal vectors is small.

In the presence of noise, we resort to signal space model in the frequency-encode direction that does not require computation of spatial derivatives. In this model, we use the constraint that the one-dimensional Fourier Transform sampled at a point on the dephasing lobe to be the output of an FIR filter whose input consists of the symmetrically reflected past samples in the rephasing lobe. The FIR filter coefficients are estimated using least squares formulation. Since the least squares solution requires knowledge of the complete echo, the data samples for the coefficient estimation are chosen from one-dimensional Fourier Transforms of the rows in a denoised version of the low-resolution image. Unlike the signal space model in the phase-encode direction, the FIR filters operate directly in the intermediate Fourier domain, without the need for a prior highpass filtering step. Also the missing information in the 2D partial kspace is simultaneously replaced during FIR filtering of each successive line in the intermediate space.

The following section discusses the theoretical background relating to each of the two types of signal space models in both phase and frequency-encode directions.

Background

In our previous work, an FIR filter derived from a complete echo with zero-phase encoding, is used for recovering missing information from a partially acquired echo sequence [16]. Such a filtering scheme is shown to achieve results comparable to other partial k-space approaches only when the noise content is less than about 0.4%. Moreover, additional noise will be introduced as a result of applying the filter on higher phase-encode lines. Increase in amplitude of phase-encode pulses leads to distortion of a normal echo shape where the signal energy is localized around the echo peak as shown in Fig. 2. The relative fraction of higher phase-encodes with peak distortion is larger for high resolution images. The suitability of a single reference filter model such as that derived from the zero-phase encode line, is therefore, limited to coarse resolution images. Consequently, application of single reference FIR filter to high resolution kspace data results only in a marginal improvement over the trivial zero-filled image.

Fig. 2 Here

A second limitation of partial kspace filling using single reference filter is the inability to compensate for phase errors. For real data, phase errors are shown to arise from main field inhomogeneities, or presence of air-tissue interface and/or susceptibility effects. This is illustrated using a simulated example in Fig. 3. A kspace is first synthesized by addition of a random phase to the real image. Imposition of conjugate symmetry constraint on one half of this kspace is seen to result in phase errors. The effect of imposition of conjugate symmetry for a real kspace is shown in Fig. 4.

Fig. 3 Here

Fig. 4 Here

Synthetic phantoms are derived for a finite resolution (N=256) by assuming a Lorentzian spectrum with $T_2'=0.3T_2$ [16]. The x and y-signals are generated for each voxel using a standard spin-echo sequence and traversing the kspace in reverse-centric manner using a total of 256 phase-encode steps. In the ideal situation, we assume that susceptibility effects and magnetic inhomogeneities are ignored. Consequently, the kspace generated will be inherently conjugate symmetric. In the examples discussed below, the images are reconstructed from missing kspace data in both frequency and phase-encode directions. In the resulting 2D partial kspace, phase errors will be introduced with the inclusion of fractional kspace lines in both directions. Fig. 5 (a)-(b) shows the application of conjugate synthesis and homodyne [9] methods for phase correction for different numbers of fractional lines, m in the frequency-encode, and q in the phase-encode direction respectively.

Fig. 5 Here

The three panels in each row represent images reconstructed from 2D partial kspace with q=10, 30, and 90 and m=45. In both conjugate synthesis and homodyne methods, we see clear cases of Gibbs ringing artefacts with repetitions of image boundary over entire FOV. The effect of artefact is reduced as the number of fractional lines is increased. Panels in (c)-(d) correspond to images reconstructed from a physical phantom acquired with a matrix size of $512 \times 512$. It is seen that the images reconstructed using conjugate synthesis from the 2D-partial kspace show ghosting artefacts in addition to Gibbs ringing artefacts, as opposed to homodyne reconstruction in which the ghosting artefacts are absent. As before, the effect of both the artefacts is reduced as the number of fractional lines is increased in either direction.

Formulation of kspace as a signal subspace model

Consider a tissue having proton density distribution $\rho(x,y)$, located within a rectangular Field-of-View (FOV) extending from $x=-Fx$ cm to $Fx$ cm and $y=-Fy$ cm to $Fy$ cm. For convenience, the FOV is divided into $N \times N$ pixels. For two-dimensional imaging, the x-direction is frequency-encoded using a gradient-field pulse of height $G_x$ T/cm, and the y-direction is phase-encoded with a short gradient-field pulse of duration λ samples, and height $G_y$ T/cm. The MRI signal originating from a single pixel location at *(x, y)* is then given by

$$S_{xy}\big[(n-m)\Delta t\big] = h\big[x, y,(n-m)\Delta t\big]\exp(\,jw(x)n\Delta t + \phi)\qquad(1)$$

where *h* is the FID signal originating at location *(x, y)*. Inverse Fourier Transform of the local spin-spectral distribution weighted by ρ(x, y), Δt is the sampling interval, echo-time $T_E=m\Delta t$,   *n=m-(Nx-1)....m+(Nx-1)*,    $w(x) = \gamma x G_x$, $\phi = k\gamma G_y \lambda \Delta t$, *k=-(Nx-1)....(Nx-1)*, and γ denotes the gyromagnetic ratio. Assuming that the FOV is such that *Fx=Fy, Gy=Gx/λ*, yields $\phi=kw(y)\Delta t$. The composite signal from all *(x,y)* locations within the FOV is given by

$$S(k,n) = \sum_x \left[ \sum_y \rho(x,y)h(x,y,p\Delta t)\exp(\,jkw(y)\Delta t) \right]\exp(\,jw(x)(n)\Delta t)$$
$$= \sum_x \varphi(x,k)\exp(\,jw(x)(n)\Delta t)$$

$$(2)$$

where *p=n-m* and $S(k,0) = \sum_x \varphi(x,k)$. Denoting the positive and negative halves of the kspace with $S^+(k,n)$ for $k \geq 0$ , and $S^-(k,n)$ for $k \leq 0$, the central line of the kspace may be considered as a superposition of two separate data sets

$$S(k,0) = \sum_x \varphi^+(x,k) + \varphi^-(x,k)\qquad(3)$$

Using the boxcar representation [14], the spatial derivative of the image can be represented as the summation of discrete impulse functions. Hence, the Fourier Transform of the spatial derivative can be modelled as a summation of finite number of sinusoids, which is linear-predictable to a certain order [15]. Using the differentiation property of Fourier Transform, the Fourier Transform of the spatial derivative $\tilde{S}^{+/-}(k,n)$ is the same as the original kspace weighted by $j2\pi k\Delta v$. The given set of *q* fractional lines of the filtered kspace in the negative phase-encoding direction, can then be used as input to predict the missing data using

$$\mathrm{Re}\left[\widetilde{S}^{+/-}(k,n)\right] = -\sum_{l=1}^{L} b_R(l)\,\mathrm{Re}\left[\widetilde{S}^{+/-}(k-l,n)\right] + e_{Rk}$$
$$\mathrm{Im}\left[\widetilde{S}^{+/-}(k,n)\right] = -\sum_{l=1}^{L} b_I(l)\,\mathrm{Im}\left[\widetilde{S}^{+/-}(k-l,n)\right] + e_{Ik} \qquad (4)$$

The predicted data is multiplied by $1/\,j2\pi k\Delta v$ to retrieve the raw unacquired data. The order of the predictor $L$, is taken to be half the size of the given data set $(q+1)/2$. If we use a fixed filter to extrapolate all the succeeding lines, the variance in the resulting filtered data points will introduce streaking artefacts and fails to completely eliminate the Gibbs ringing artefacts. As a first step towards elimination of such artefacts, we propose an iterative approach for extrapolating higher phase-encode data in the negative direction using filters with successively larger prediction orders. This is illustrated in a block schematic form in Fig. 6.

Fig. 6 Here

At each prediction step, the unbiased autocorrelation of the filtered kspace is computed and input to the Levinson algorithm [17], for estimating the filter coefficients. Using the filter coefficients, the next sample in the higher phase-encode line is predicted using Eq. (4). In the iterative approach, this estimated sample is reused to form the new input to the Levinson algorithm. In the first prediction step, the error output of the predictor is realized as a regular process of the Wold's decomposition [18]. However, by inclusion of the predicted sample into the data set for the succeeding prediction step, the error generated by the predictor would now consist of an additional component that cannot be modelled as a regular process. As the algorithm is iterated through larger number of steps, the non-regular component of prediction error will introduce artefacts in the reconstructed image. In this process, if the filter order is maintained constant, then imposing the orthogonality condition can compensate the non-regular component of the error. However, as the data size increases, the usage of a fixed filter size introduces additional bias into the predicted estimates. Therefore, with a successive increment of the filter order in each step, the compensation can only be achieved using subspace approximation methods.

The dataset consisting of the positive phase-encode steps at a given time-point is first distributed into *L+1* dimension subspace by casting the signal samples in the form of a prediction matrix [19]

$$\tilde{\mathbf{S}}^+[n] = \left[\tilde{\mathbf{S}}_L^+[n], \tilde{\mathbf{S}}_{L-1}^+[n], \cdots\cdots \tilde{\mathbf{S}}_1^+[n], \tilde{\mathbf{S}}_0^+[n]\right] \tag{5}$$

where $\tilde{\mathbf{S}}_i^+[n] = \left[\tilde{S}^+(L+1-i,n), \tilde{S}^+(L+2-i,n), \cdots\cdots, \tilde{S}^+(Nx+i-1,n)\right]^T$. By application of Gram-Schmidt orthogonalization procedure to L+1 dimensional subspace $\tilde{\mathbf{S}}^+[n]$ , we obtain

$$\tilde{\mathbf{S}}^+[n] = \tilde{\mathbf{V}}^+[n]\tilde{\mathbf{B}}^+[n] \tag{6}$$

where $\tilde{\mathbf{v}}^+[n]$ represents the orthogonal basis vectors $\left[\tilde{\mathbf{V}}_L^+[n], \tilde{\mathbf{V}}_{L-1}^+[n], \cdots\cdots \tilde{\mathbf{V}}_1^+[n], \tilde{\mathbf{V}}_0^+[n]\right]$ spanning the *L+1* dimension subspace with $\left(\tilde{\mathbf{v}}^+[n]\right)\left(\tilde{\mathbf{v}}^+[n]\right)^{\mathbf{H}} = diag(\theta_1, \theta_2, \cdots\theta_{L+1}) = \tilde{\mathbf{\Theta}}^+$, and $\tilde{\mathbf{B}}^+[n]$ is an upper triangular matrix with unit diagonal elements. In the current approach, the closest approximation of the iteratively predicted data from the acquired set of fractional lines in the negative phase-encoding direction to the subspace in Eq. (6) is calculated using the projection theorem [20]. The projected data is then lowpass filtered by multiplying with $1/j2\pi k\Delta v$ to retrieve the missing kspace data.

A geometrical interpretation of the error compensation procedure is illustrated with the help of a vectorial representation of the component subspaces in Fig. 7. Fig. 7(a) shows sample signal vectors in both data sets for each step in the prediction algorithm. Since the true signal samples $S^+(k,n)$ are known *apriori* for the signal subspace in the positive phase-encode direction, the inclusion of predicted sample into each new input to the Levinson block, may be replaced using the corresponding value of the true signal sample. Consequently, with each increment in the subspace dimension in the iterative process, the error vector will be orthogonal to the signal vector. For the r'th iteration, the signal and error vectors are represented using $S_{k+r}^+$ and $e_{k+r}^+$. In the negative phase encoding direction, the orthogonality is maintained for the first iteration. For successive iterations, the predicted output is no longer orthogonal to the error vector due to the inclusion of the previous predicted sample into the data input to the Levinson algorithm. In the vectorial representation, the predicted and true signal vectors in Fig. 7(b) are denoted by $\hat{s}_{k+r}^-$ and $s_{k+r}^-$ respectively. The errors before and after compensation are given by

$$E_{Tk}^{-2} = S_{k+r}^{-2} + \hat{S}_{k+r}^{-2} - 2S_{k+r}^-\hat{S}_{k+r}^-\cos(\theta)$$
$$E_{Ck}^{-2} = S_{k+r}^{-2} + \hat{S}_{k+r}^{-2}\cos^2(\theta + \Delta\theta) - 2S_{k+r}^-\hat{S}_{k+r}^-\cos(\theta + \Delta\theta)\cos(\Delta\theta) \tag{7}$$

In order for the error compensation to work, it is required to satisfy the condition $\|E_{Ck}\| << \|E_{Tk}\|$. From Eq. (7), approximating $\Delta\theta$ to be sufficiently small, the necessary and sufficient condition for error compensation can be shown to be $\|\hat{s}_{k+r}^- + \hat{s}_{k+r}^-\cos(\theta)\| << 2\|S_{k+r}^-\|$. By limiting the number of iterations in the prediction algorithm, the phase angle $\theta$ between the predicted and true signal vectors will be small. For limited number of iterations, the compensation criterion simplifies to $\|\hat{s}_{k+r}^-\| << \|S_{k+r}^-\|$. Thus the compensation method will be effective only in cases where the norm of the predicted signal vector is less than that of the original signal vector. Also, in the presence of excessive levels of noise in both the x and y coils, the angle $\Delta\theta$ between the positive and negative phase-encode signal subspace vectors cannot be neglected in Eq. (7). Consequently, the compensation method will not generate the desired performance. In the succeeding section, we present an FIR model derived from the one-dimensional Fourier Transform in the x-direction of the denoised low-resolution image. This filter can be used to directly fill the missing kspace information in the frequency encode direction. In comparison to earlier model based approaches for kspace regression [14], the proposed FIR model operates directly in the intermediate space, without the need for a prior highpass filtering step. This alleviates artefact generation in the image reconstructed from noisy partial kspace data.

Fig. 7 Here

Partial-Echo Prediction using Least-Squares FIR filter

Rewriting Eq. (2) in the form

$$S(k, n) = \sum_y \psi(y, n) \exp(jkw(y)\Delta t) \qquad (8)$$

where $\psi(y_0, n)$ represents the one-dimensional Fourier Transform of the object function $\rho(x, y_0)$. However, since we are not provided with apriori information about the finer details of the object function, we start off with a low-resolution approximation. This is obtained using a denoised version of the low-resolution image reconstructed from the incomplete kspace, with m fractional lines in the frequency-encode, and q fractional lines in the negative phase-encode directions respectively. The denoising is performed using a Non-Local Means (NLM) filter [21]. Following the notations in [14], the low-resolution object function can be represented as the summation of a finite number of boxcar functions with edge locations $\varepsilon_1 < \varepsilon_2 < \dots \varepsilon_M$. Using the above representation, the one-dimensional Fourier Transform can be expressed as

$$\psi(y, n) = \sum_{i=1}^M \alpha_i \exp\left(-j \frac{\pi n(\varepsilon_i + \varepsilon_{i+1})}{N}\right) \frac{\sin\left[\frac{\pi n}{N}(\varepsilon_{i+1} - \varepsilon_i)\right]}{\sin\left[\frac{\pi n}{N}\right]} \qquad (9)$$

where $\alpha_i$ is the amplitude of the $i$'th boxcar function. The one-dimensional DFT will have a magnitude spectrum that is quasi-symmetric about $n=0$. In the proposed model, we use the constraint that a spectral sample for a given value of $n > 0$ is the output of an FIR filter consisting of the symmetrically reflected past samples. This is mathematically represented using

$$\psi(y,n) = -\sum_{i=1}^{Q} a(y,i)\psi\big(y,-(n-i)\big) \qquad (10)$$

Substitution of $\psi(y,n)$ into Eq. (8) yields

$$S(k,n) = -\sum_{i=1}^{Q}\sum_{y} a(y,i)\psi(y,-(n-i))\exp(jkw(y)\Delta t) \qquad (11)$$

Assuming the filter order is $Q$, the filter coefficients can be estimated using the matrix equation

$$\Psi_{y,Q} = -\Lambda_{y,Q}\bar{\mathbf{a}}(y) \qquad (12)$$

where $\Psi_{y,Q} = \begin{bmatrix} \psi(y,Q) \\ \psi(y,Q+1) \\ \vdots \\ \psi(y,Nx-1) \\ \psi^{*}(y,0) \\ \psi^{*}(y,1) \\ \vdots \\ \psi^{*}(y,Nx-Q-1) \end{bmatrix}$, and

$$\Lambda_{y,Q} = \begin{bmatrix} \psi(y,-(Q-1)) & \psi(y,-(Q-2)) & \cdots & \psi(y,0) \\ \psi(y,-(Q)) & \psi(y,-(Q-1)) & \cdots & \psi(y,-1) \\ \vdots & \vdots & \vdots & \vdots \\ \psi(y,-(Nx-2)) & \psi(y,-(Nx-3)) & \ldots & \psi(y,-(Nx-Q-1)) \\ \psi^{*}(y,-1) & \psi^{*}(y,-2) & \ldots & \psi^{*}(y,-(Q)) \\ \psi^{*}(y,-2) & \psi^{*}(y,-3) & \cdots & \psi^{*}(y,-(Q+1)) \\ \vdots & \vdots & \cdots & \vdots \\ \psi^{*}(y,-(Nx-Q)) & \psi^{*}(y,-(Nx-Q-1)) & \cdots & \psi^{*}(y,-(Nx-1)) \end{bmatrix}$$

From Eq. (12), the filter coefficients $\bar{\mathbf{a}}(y) = [a(y,1), a(y,2) \cdots a(y,Q)]^T$ are estimated using the least-squares solution. A schematic diagram illustrating the steps involved in the estimation of missing kspace data is shown in Fig. 8.

Fig. 8 Here

Results

Image reconstruction using various kspace extrapolation techniques is compared using both simulated and real data sets. The real data set is acquired from raw MR data, on a 1.5T twin speed clinical scanner (GE Healthcare, Milwaukee, USA), using a 2D single-echo Carr-Purcell sequence (matrix size of 512 × 512, slice thickness of 3mm, TR=500ms, TE=22ms, +/-15.63 kHz readout bandwidth, and 18cm field-of-view). Volume imaging is simulated using a set of 10 axial slices (numbered 1 to 10) of resolution 255 × 255. Complete acquisition with $T_E$=100 msec, and $90^0$ RF pulses along with slice selection applied at an interval of 150 msec results in a $T_{Rep}$ of 1500 msec. For partial acquisition, the $90^0$ RF pulses are applied earlier at an interval of 112 msec ($\mu_s$=1.1176). For the same $T_{Rep}$ of 1500 msec, partial-echo sequence enables acquisition of additional 3 slices.

Fig. 9(a) shows a 512 × 512 image obtained by applying homodyne phase-correction method to the image reconstructed by Fourier transforming a zero-filled partial kspace with $q$=65 fractional lines in the phase-encode direction, and $m$=65 fractional lines in the frequency–encode direction respectively. Fig. 9(b) shows the image reconstructed from partial k-space using iterated prediction in the phase encode direction. The iteration is started with $q$=35 fractional lines in the phase encode direction, with the additional 30 lines extrapolated using the algorithm. The image resulting from signal space projection on the extrapolated lines is shown in Fig. 9(c). As observed from the zoomed images shown in bottom panels, the improved efficiency of subspace projection is evident from the similar performance to homodyne phase-correction, but using half the number of acquired fractional lines.

Fig. 9 Here

Fig. 10 illustrates the results for partial echo prediction using least squares FIR filters. The input image is obtained by Fourier transforming a partial kspace synthesized using spin-echo simulation. The details of simulation procedure are outlined in Appendix A-1. The partial kspace is obtained using a phase-encode coverage ranging from $-qGy$ to $(Nx\text{-}1)Gy$ and the echoes truncated during the dephasing lobe. For a resolution of 255 × 255, $q$=10 fractional lines in the phase-encode, and m=45 fractional columns in the frequency-encode directions are chosen for image reconstruction. The phase-errors due to the fractional lines are initially compensated using the homodyne algorithm. However, truncation artifacts are still found to be present in the resulting image shown in panel (a). As a first step for artifact suppression, this image is filtered using a Non-Local Means (NLM) filter [21]. A search window radius of $t$=5, and a similarity window radius of $f$=1 are chosen to perform NLM filtering operation. The resulting low-resolution image is shown in panel (b). Model creation is accomplished by Fourier-transforming each row of the low-resolution image, followed by formulation of the matrices outlined in Eq. (12). This leads to the determination of the FIR filter coefficients corresponding to each echo in the intermediate domain. The bank of filters is then used to estimate the missing information in the intermediate kspace of the input image. The image reconstructed from the extrapolated kspace is shown in panel-(c). It is observed that the artifacts due to truncation are now completely eliminated.

Fig.10 Here

## Discussion and Summary

Besides artefact reduction, reconstruction using FIR filters in the frequency-encode direction is seen to result in a 20% increase in the CNR. The increase in CNR against the number of fractional lines is plotted in Fig. 11.

Fig. 11 Here

Of the two methods for extrapolating kspace data, the one using prediction in the phase-encode direction is very sensitive to noise. Since the frequencies present in the simulated received signal for the synthetic data, extend beyond ROI in the readout direction, there is an excess of aliased noise. The effect of this noise is clearly evident in Fig. 12 showing images reconstructed using all four methods, viz. 1) linear prediction applied to frequency-weighted kspace, 2) iterated prediction in phase-encode direction, 3) projection onto signal space, and 4) FIR filters in the frequency-encode direction.

Fig. 12 Here

The percentage reduction in truncation artefacts is quantified by computing errors in the edge images obtained using canny filters. Plots of percentage error versus number of fractional lines, shown in Fig. 13 provide a means for comparing the performance in eliminating these artefacts. In summary, extrapolation using FIR filters in the frequency-encode direction provide the best performance in respect of artefact reduction and improvement in CNR.

Fig. 13 Here

## Acknowledgement

The authors wish to thank the Department of Science and Technology (DST) of India for scholarship and operating funds,

Appendix A-1 (Numerical Simulation)

For numerical simulation, a brain axial slice phantom is constructed using the proton-density ($\rho$), $T_1$, and $T_2$ values for a static field $B_0$=1.5T. The values chosen for simulation are shown in Table-1.

| | White Matter | Gray Matter | CSF |
|---|---|---|---|
| $\rho$ | 0.65 | 0.8 | 0.9 |
| $T_1$(ms) | 650 | 950 | 4000 |
| $T_2$(ms) | 80 | 100 | 2000 |

Table-1: Tissue parameters for 1.5T

For simplicity, the slice thickness is assumed to be zero, so that the phantom consists a finite number of 255 x 255 locations in the axial plane. Each location is assumed to consist of a finite number of spins with a Lorentzian off-resonance frequency distribution. The simulation is carried out at three consecutive levels. In the first level, the temporal progression of the magnetization vector is computed for each slice location. This requires knowledge of $N_f$ (The number of spins), F (The Bandwidth), $\Delta t$ (The sampling time (ms) ), $\lambda$ (Number of time samples within the duration of the phase-encoding pulse), $M_{org}$ (Initial magnetization vector), $\alpha$ (Flip angle), $T_E$ (echo-time (ms) ), $T_{rep}$ ( Repetition Interval (ms) ), and the gradient amplitudes $G_x$, $G_y$ (T/cm). For each spin, the magnetization vector upon application of an $\alpha$-degree RF pulse about the y-direction in the axial slice is given by

$$\begin{bmatrix} M_{xi}(0+) \\ M_{yi}(0+) \\ M_{zi}(0+) \end{bmatrix} = \begin{bmatrix} \cos\alpha & 0 & \sin\alpha \\ 0 & 1 & 0 \\ -\sin\alpha & 0 & \cos\alpha \end{bmatrix} M_{org(i)} \qquad (13)$$

For the initial application of RF pulse, the value of $M_{org(i)}$ is determined using the proton density ($\rho$), the spin off-resonance frequency $\omega_i$, and the spectral amplitude $H(\omega_i)$ as

$$M_{org(i)} = \rho(x,y).H(\omega_i)\Delta\omega \qquad (14)$$

where $\Delta\omega$ is the sampling interval of the discretized spin-spectral distribution. The temporal evolution of magnetization vector over a duration $t'$ is then calculated using propagation matrices $A(t')$ and $B(t')$ (Kwan et al., 1999, Hargreaves: *http://mrsrl.stanford.edu/~brian/bloch*), corresponding to the time periods representative of free-precession ($A_0(t')/B_0(t')$), duration of frequency-encoding ($A_x(t')$ /$B_x(t')$), or phase-encoding ($A_y(t')/B_y(t')$). From the Bloch equations, the time progression of the magnetization vector is given by

$$\begin{bmatrix} M_x(t+t') \\ M_y(t+t') \\ M_z(t+t') \end{bmatrix} = A(t') \begin{bmatrix} M_x(t) \\ M_y(t) \\ M_z(t) \end{bmatrix} + B(t') \qquad (15)$$

where

$$A(t') = \begin{bmatrix} \exp(-t'/T_2(x,y)) & 0 & 0 \\ 0 & \exp(-t'/T_2(x,y)) & 0 \\ 0 & 0 & \exp(-t'/T_1(x,y)) \end{bmatrix} \begin{bmatrix} \cos\phi & -\sin\phi & 0 \\ \sin\phi & \cos\phi & 0 \\ 0 & 0 & 1 \end{bmatrix}$$

and

$$B(t') = \begin{bmatrix} 0 \\ 0 \\ 1 - \exp(-t'/T_1(x,y)) \end{bmatrix}.$$

For free-precession, the accumulated phase $\phi_{i0} = \omega_i t'/10^3$. During the periods of frequency and phase-encoding, the value of $\phi_i$ is appropriately modified according to $\phi_{ix} = (\omega_i + w(x)) t'/10^3$ and $\phi_{iy} = (\omega_i + kw(y)) t'/10^3$, for $k=Nx-1$ to 0 in the reverse-centric scheme for phase encode steps.

# Figures

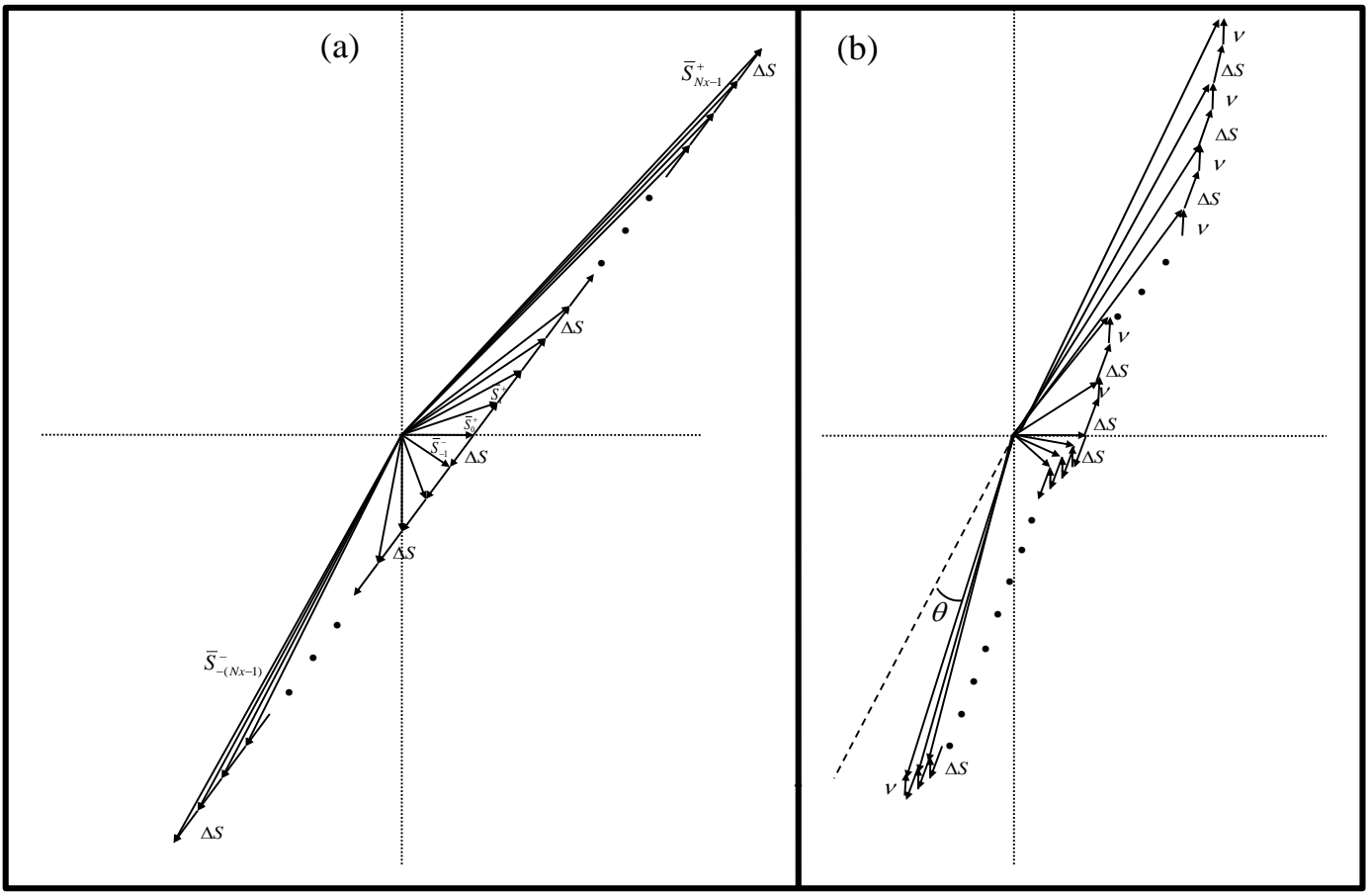

**Fig 1: Evolution of signal vectors in the phase encode direction (a) in absence of noise, (b) in presence of noise.**

(a)

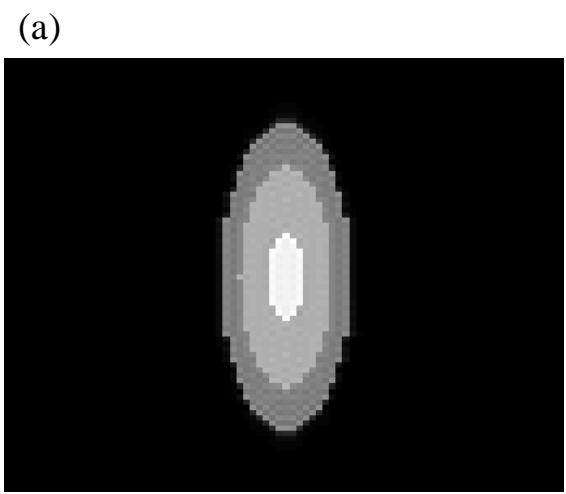

(b)

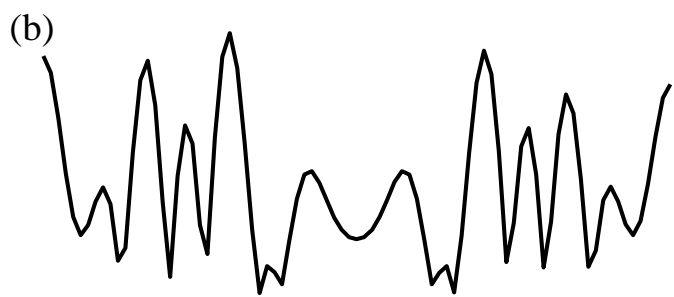

(c)

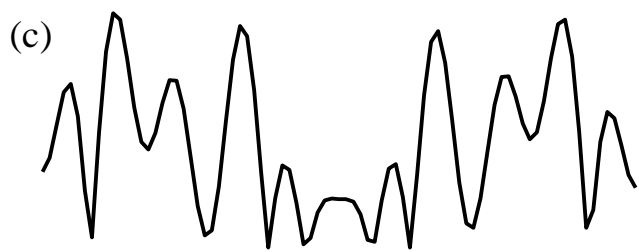

(d)

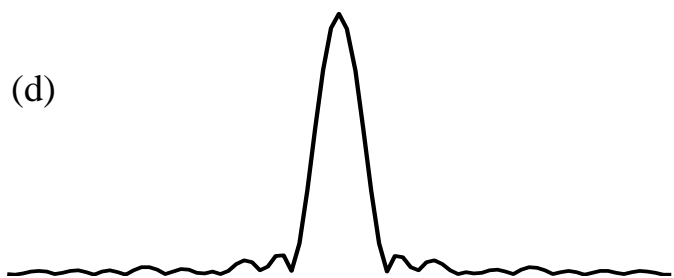

**Fig 2: Peak variations of elliptical image (a) Image whose peak is studied, (b) Top most phase encode line of k-space, (c) Second top most phase encode line of k-space, (d) Zero phase encode line of k-space.**

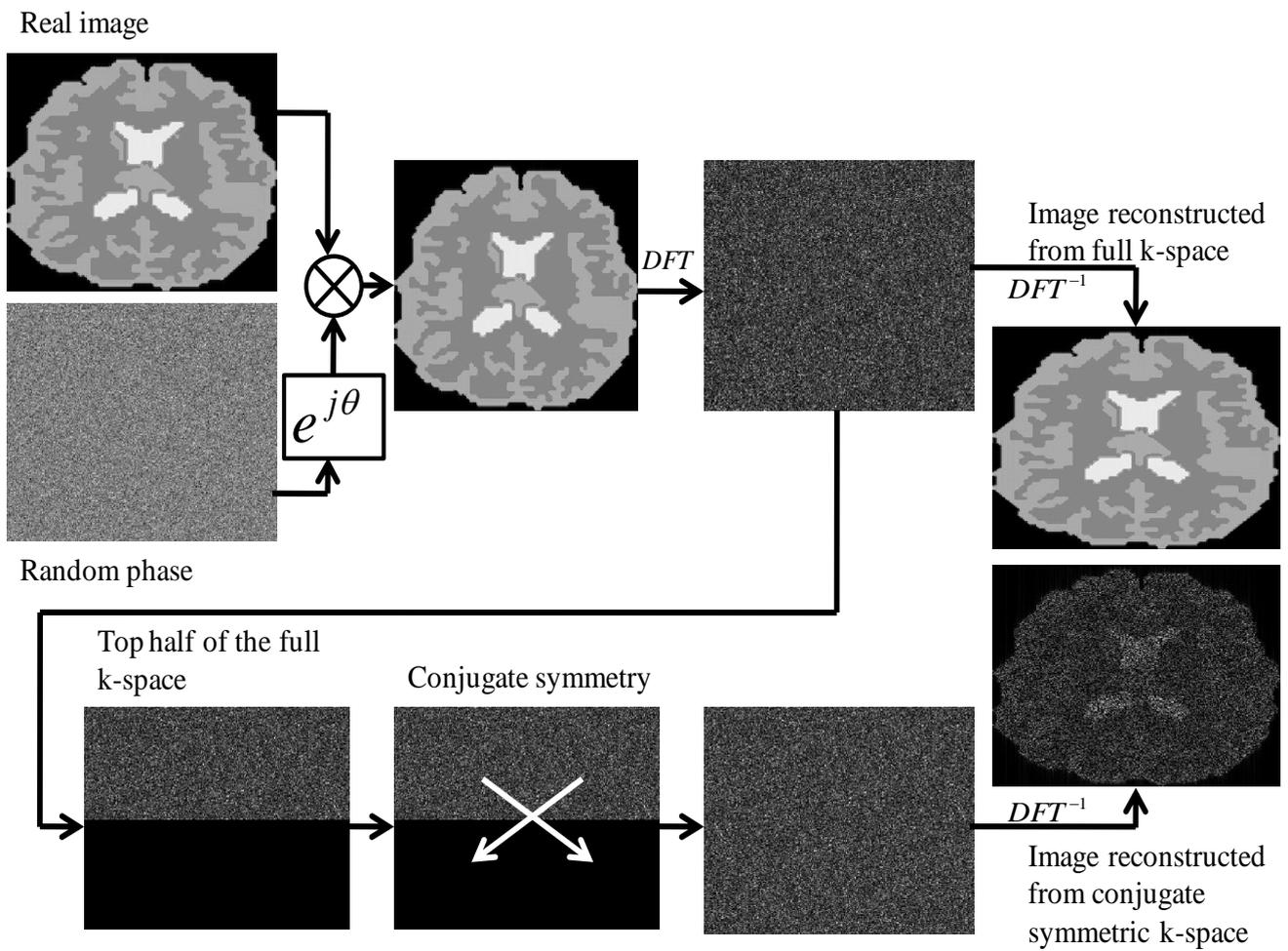

Real image

Random phase

$e^{j\theta}$

DFT

Image reconstructed
from full k-space

$DFT^{-1}$

Top half of the full
k-space

Conjugate symmetry

$DFT^{-1}$

Image reconstructed
from conjugate
symmetric k-space

**Fig 3: Illustration of generation of phase errors**.

(a) 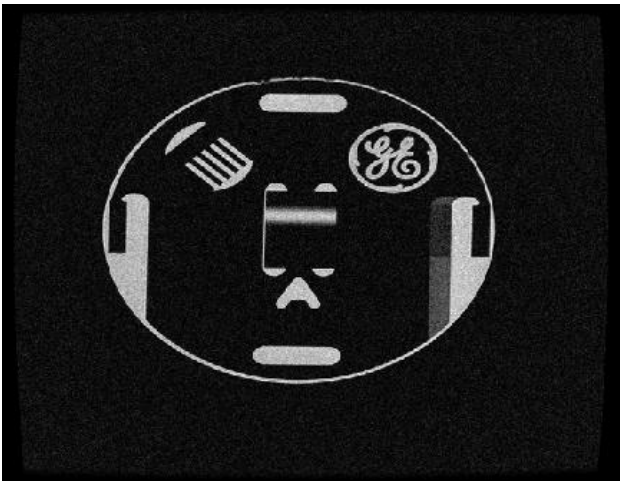 (b) 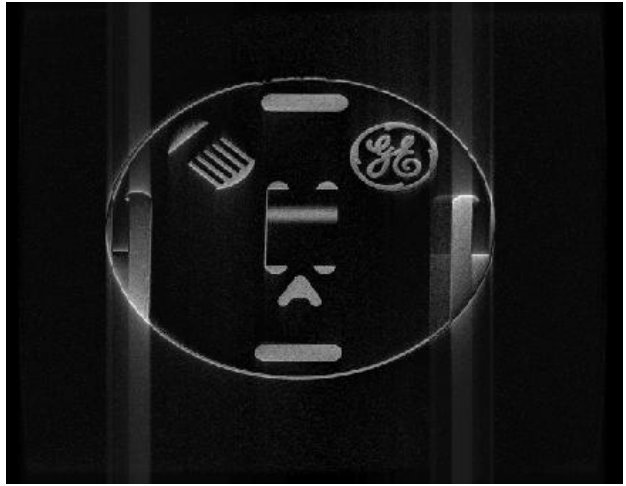

**Fig 4: (a) Image reconstructed using complete k-space data, (b) Image reconstructed using the top half of the k-space, by enforcing conjugate symmetry constraint.**

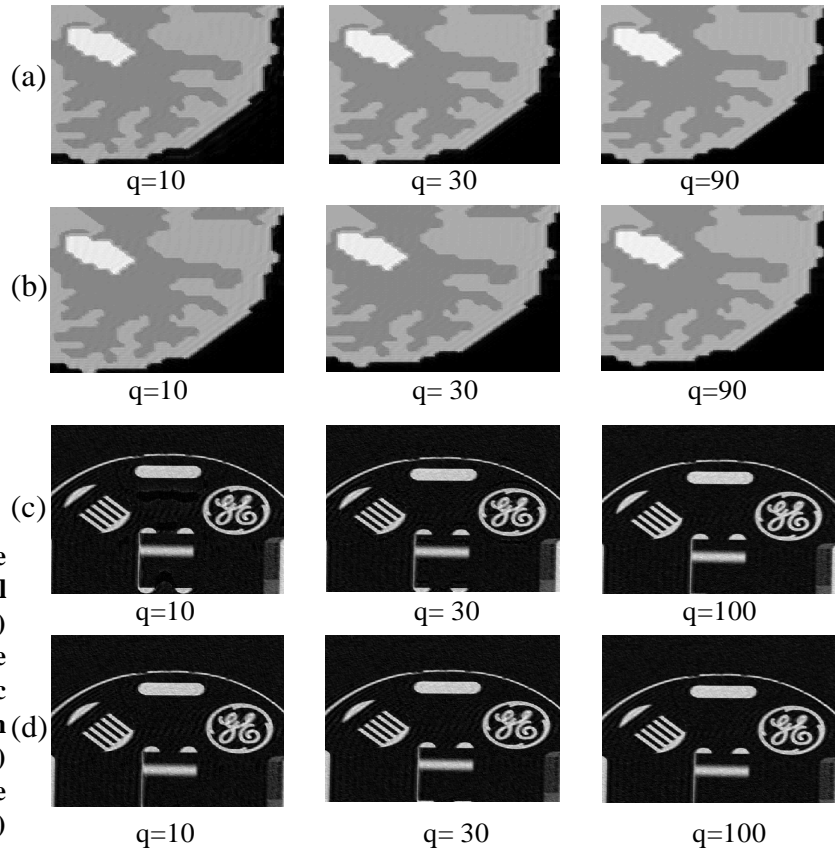

(a)

q=10    q= 30    q=90

(b)

q=10    q= 30    q=90

(c)

q=10    q= 30    q=100

(d)

q=10    q= 30    q=100

**Fig 5: Artifacts in image reconstructed using conventional phase correction techniques. (a) Phase correction using conjugate synthesis applied to synthetic data, (b) homodyne algorithm applied to synthetic data, (c) phase correction using conjugate synthesis applied to real data, (d) homodyne algorithm applied to real data.**

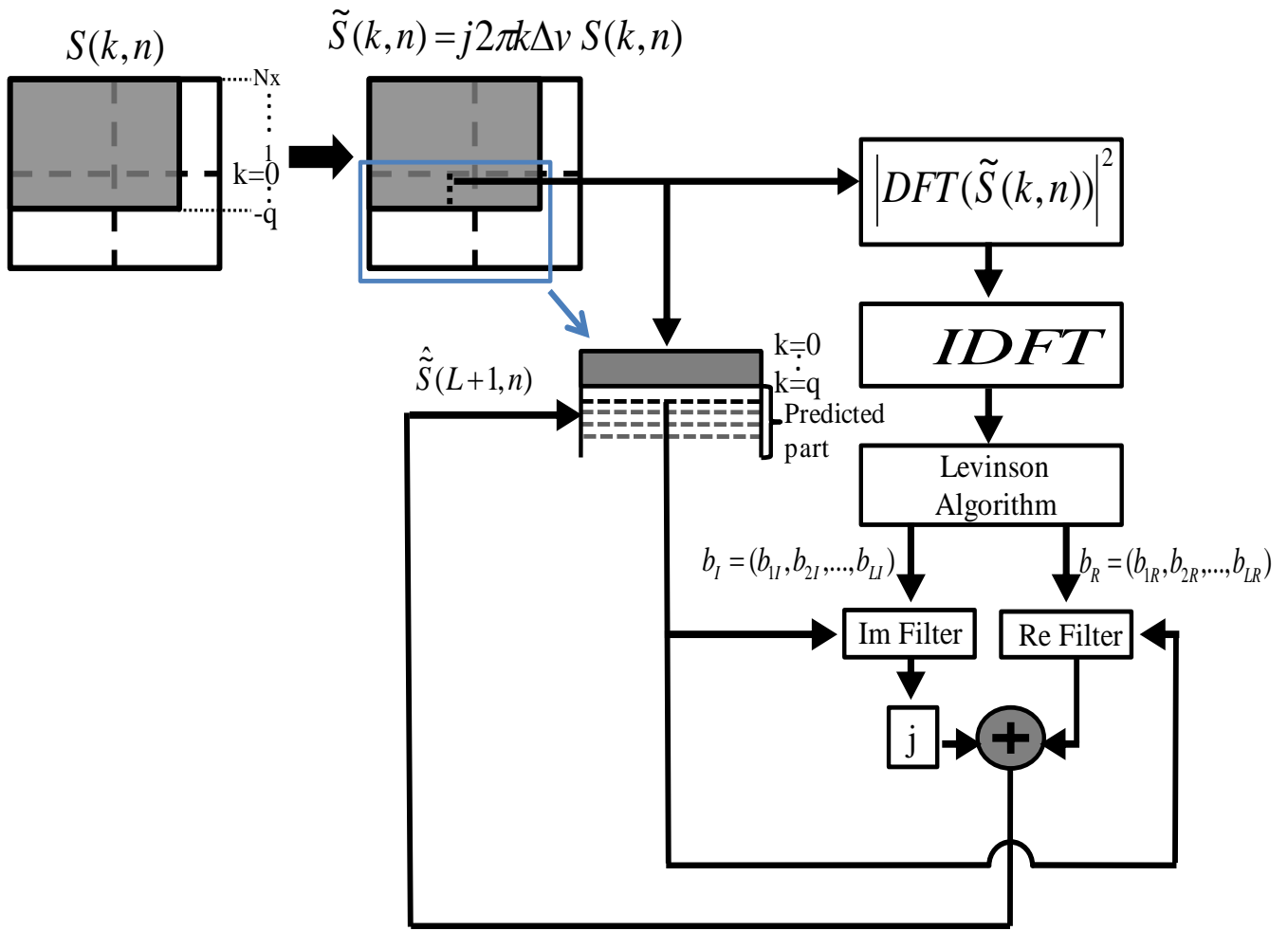

**Fig 6: Block diagram for partial k-space reconstruction in the phase encoding direction using iterated prediction.**

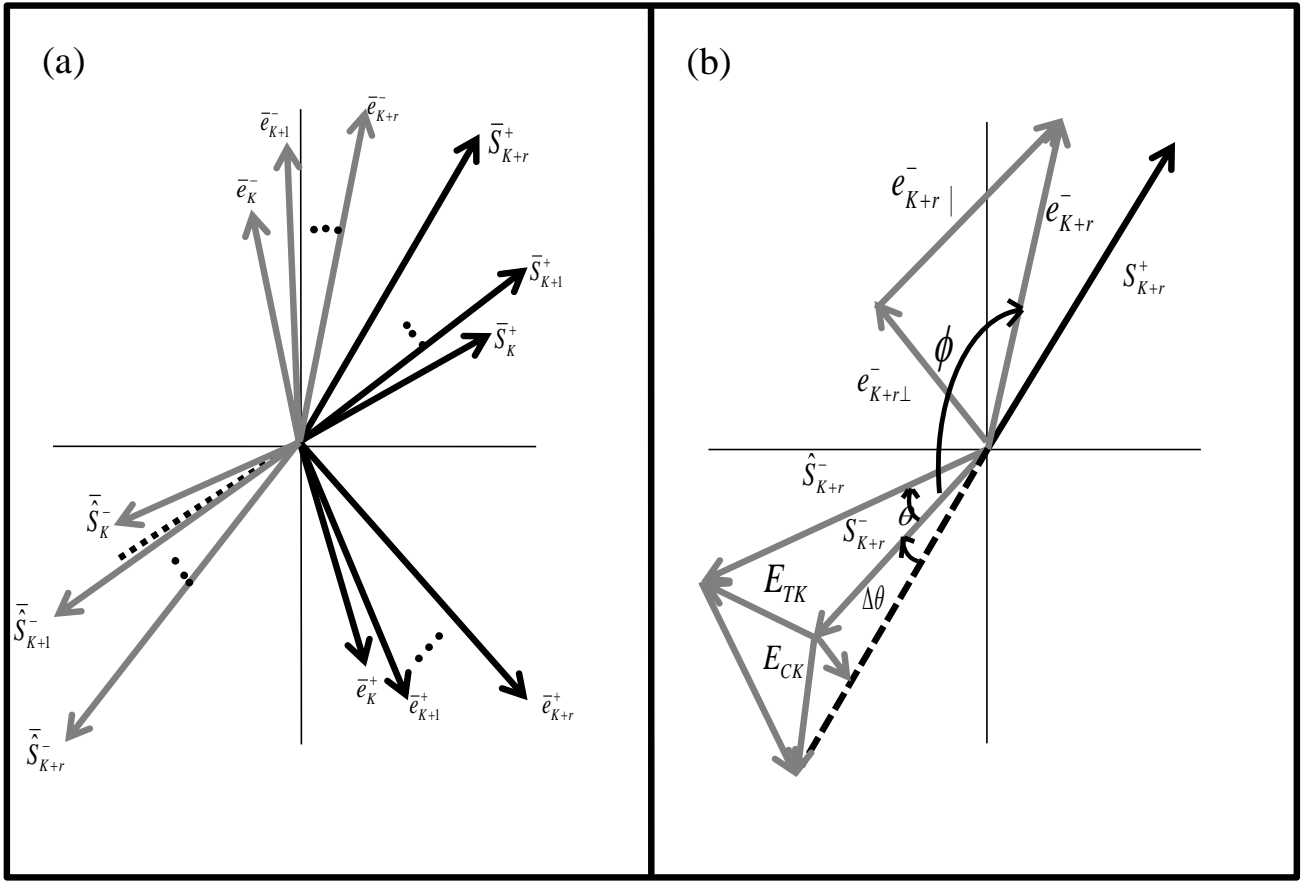

**Fig 7: Vectorial representation of projection theorem.**

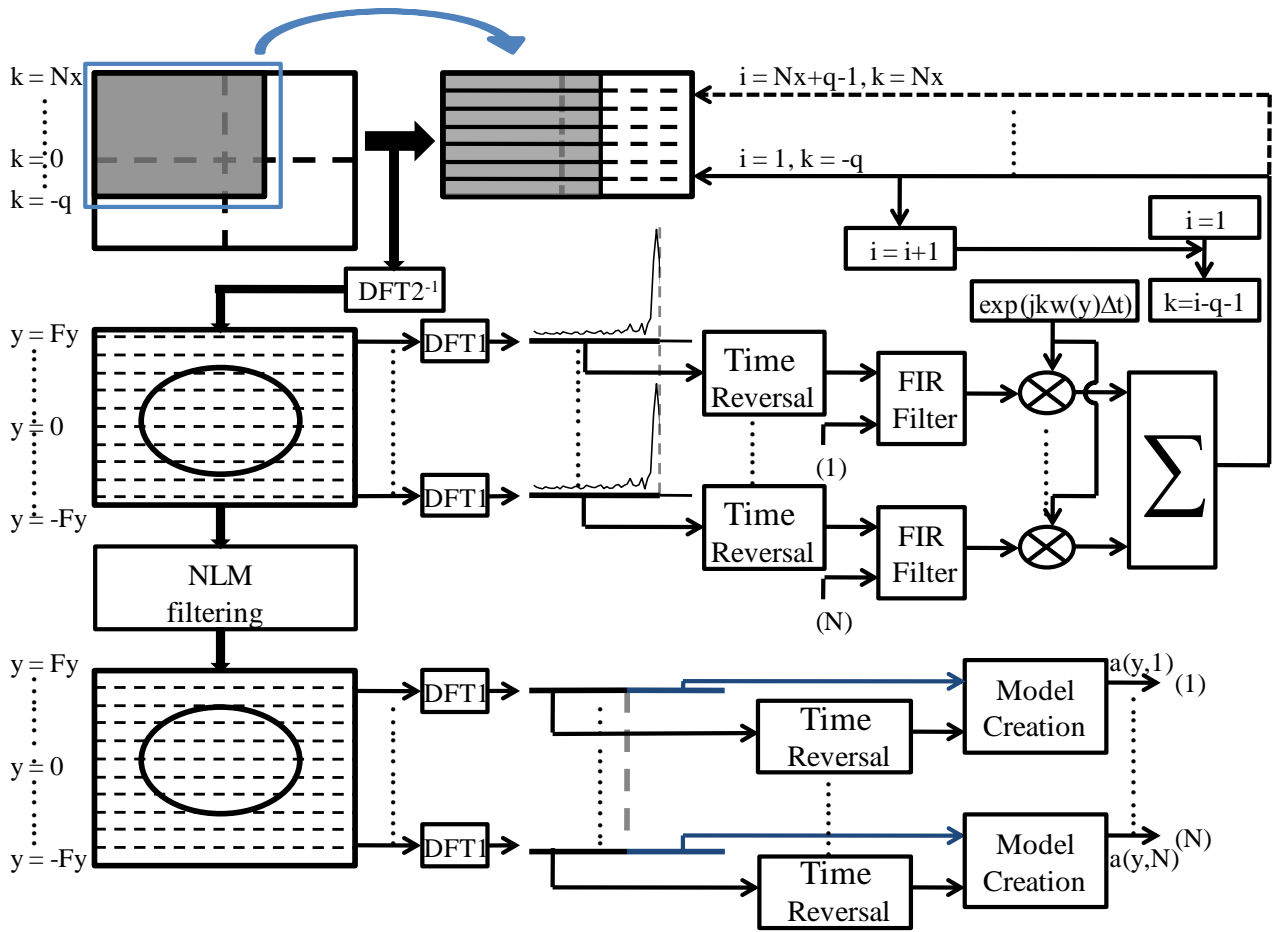

**Fig 8: FIR filter in the frequency-encode direction.**

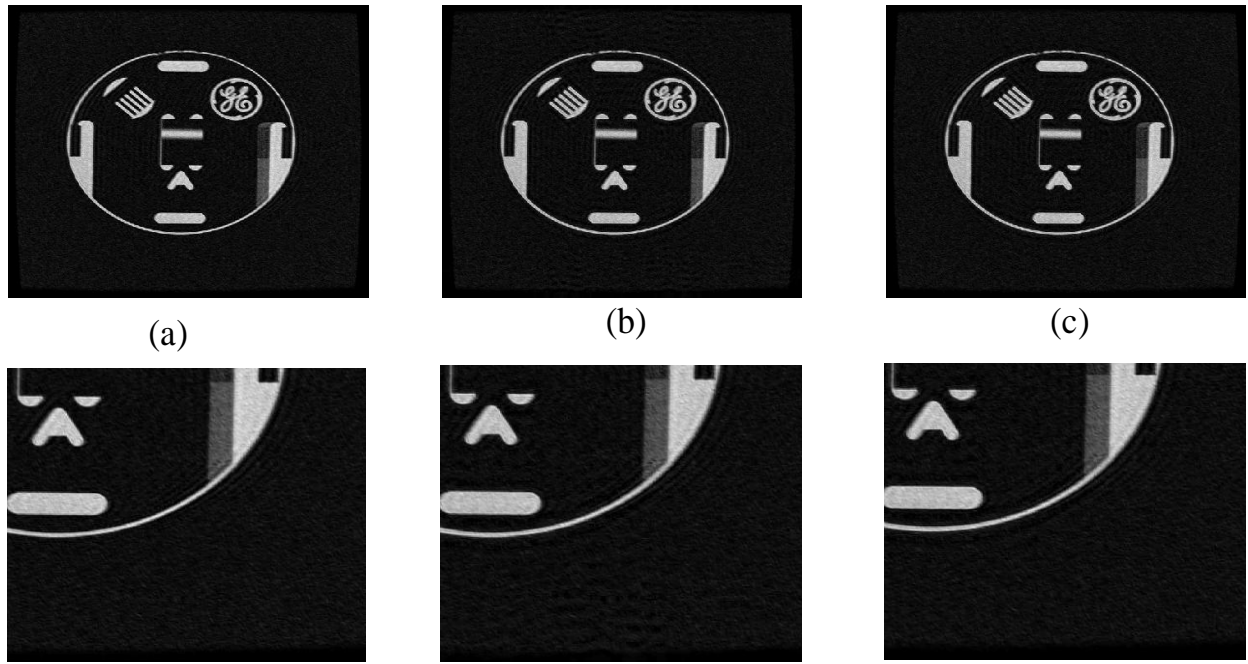

(a)  (b)  (c)

**Fig 9: Image reconstruction using iterated prediction employing lesser (q=35) number of fractional lines is shown to be comparable to that of homodyne algorithm using double the number of fractional lines (q=65). This shows improved efficiency of the proposed method. (a) Image reconstructed from zero filled partial k-space using conventional homodyne algorithm with q=65 fractional lines in the phase encode direction, (b) Image reconstructed from partial k-space using iterated prediction in the phase encode direction with q=35 fractional lines. The iteration is started with q=35 fractional lines in the phase encode direction with an additional 30 lines predicted using the algorithm, (c) Image reconstructed from predicted k-space after phase compensation using subspace projection.**

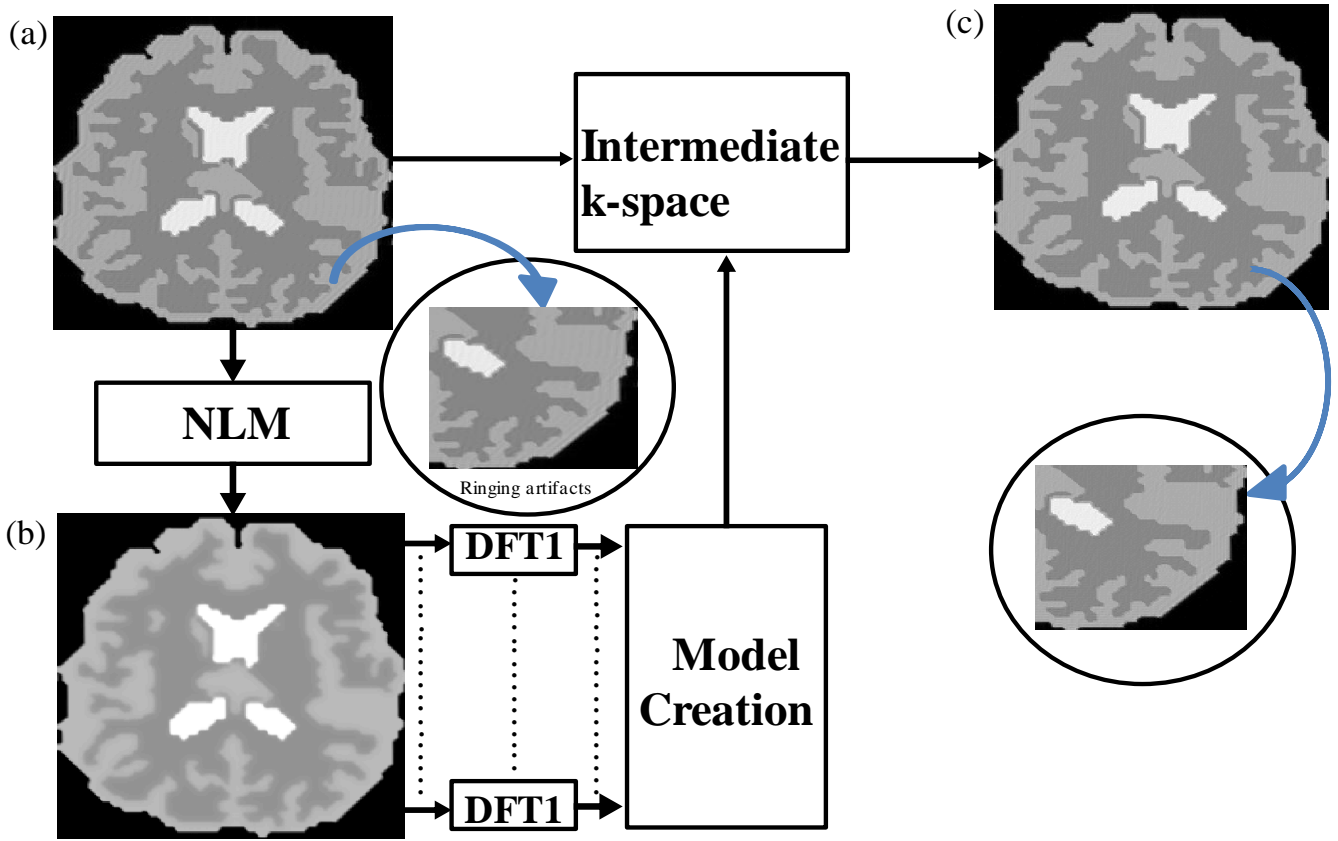

Fig 10: Application of least squares FIR filters to simulated data of size 255 x 255. (a) Image reconstructed from zero filled partial k-space using homodyne algorithm, (b) Denoised image, (c) Image reconstructed using FIR filters in the frequency encoding direction

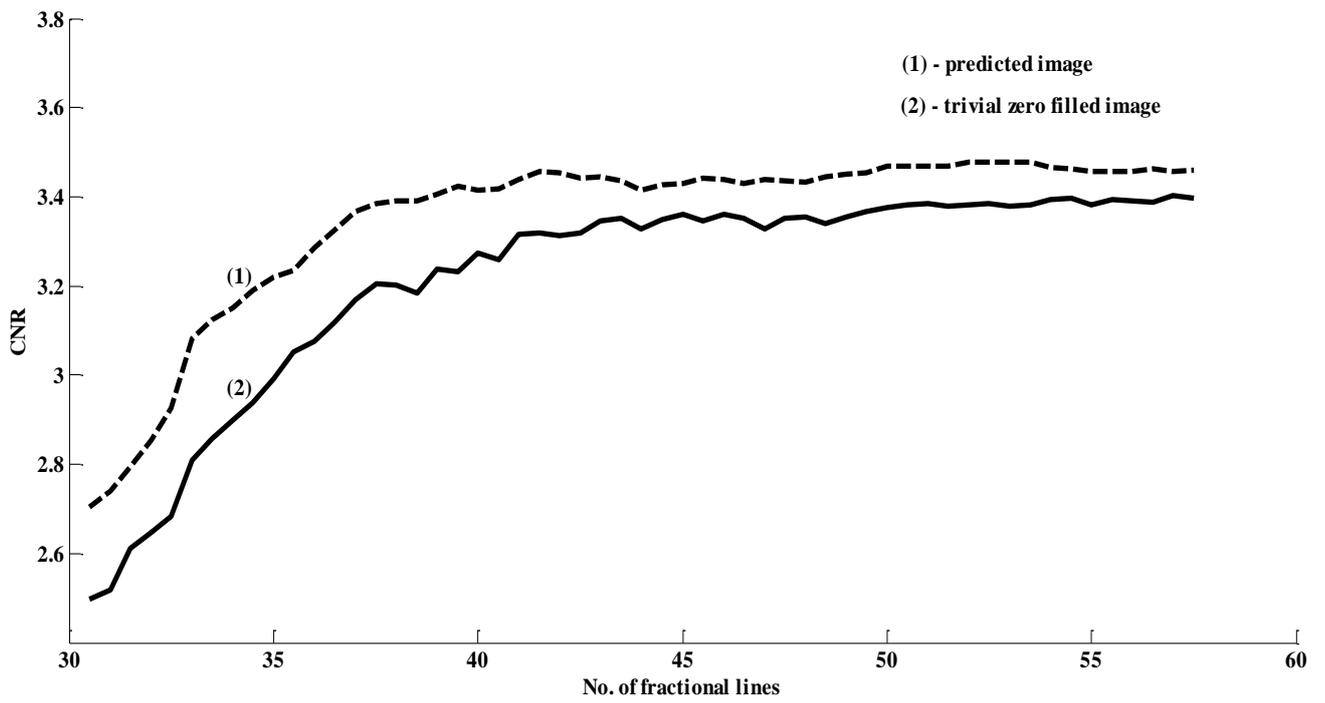

**Fig 11: CNR versus no. of fractional lines for predicted and trivial zero filled images.**

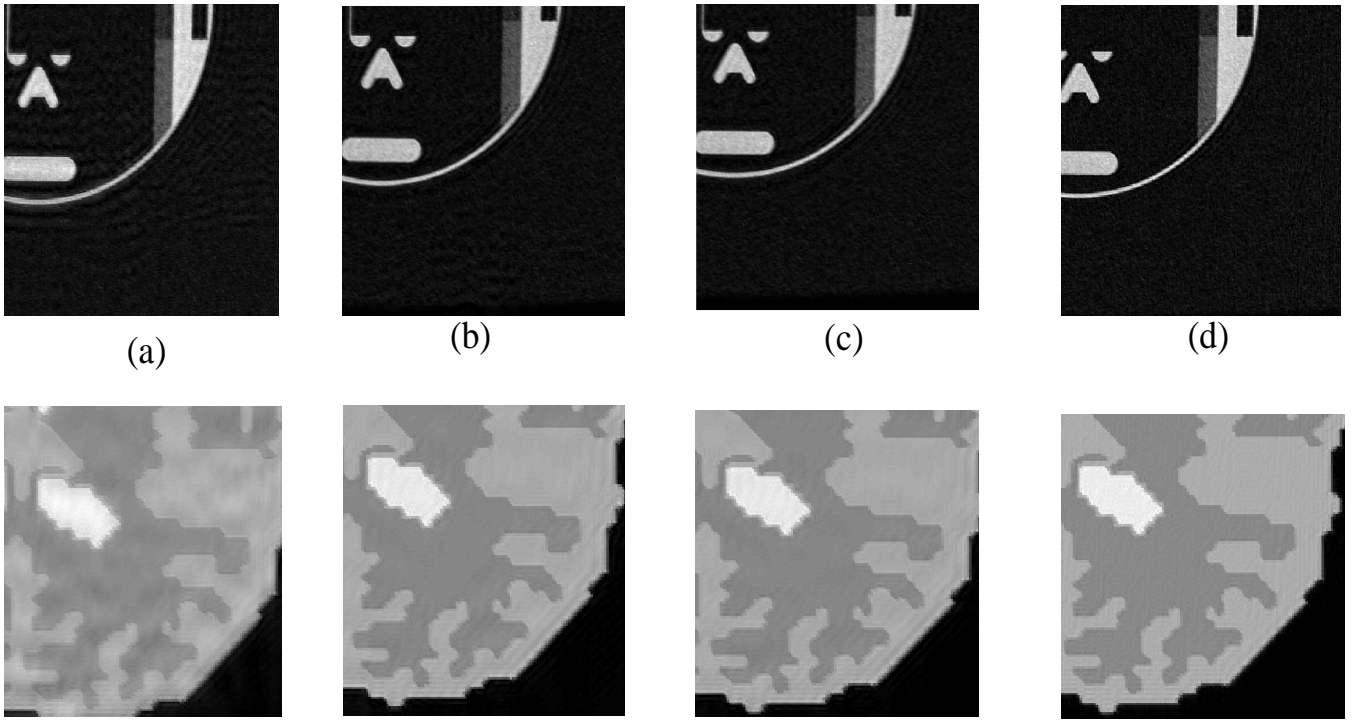

(a)   (b)   (c)   (d)

**Fig 12:** **Performance comparison of image reconstruction using various kspace extrapolation methods. (a) Extrapolation in the phase-encode direction using linear prediction applied to the frequency-weighted kspace, (b) iterated prediction, (c) projection into signal subspace, (d) FIR filter in the frequency-encode direction.**

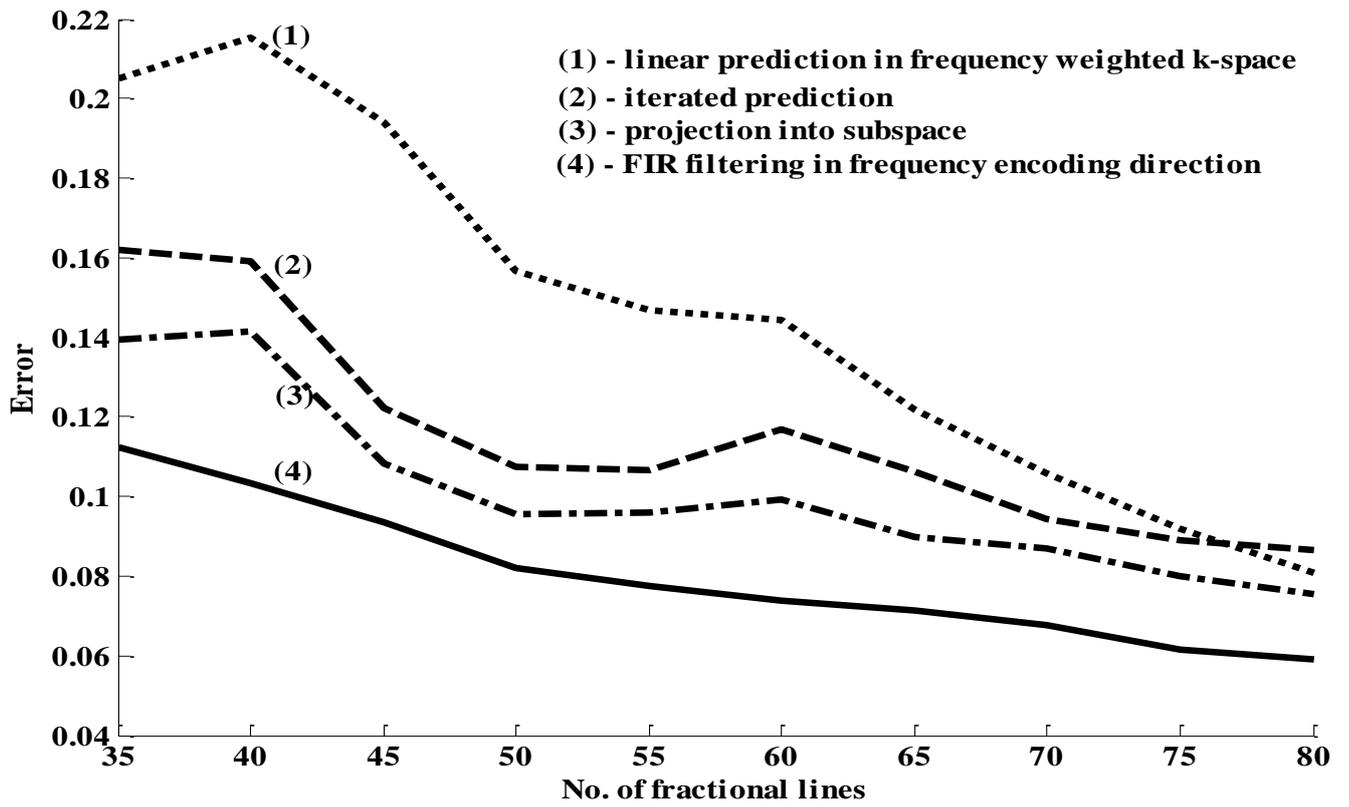

Fig 13: Percentage reduction in ringing artifacts for images reconstructed using four different types of kspace extrapolation methods.